%% file: main.tex
\documentclass[10pt,twocolumn,letterpaper]{article}

\usepackage{cvpr}              

\input{preamble}

\def\paperID{21821} 
\def\confName{CVPR}
\def\confYear{2026}

\title{SToRe3D: Sparse Token Relevance in ViTs for Efficient \\ Multi-View 3D Object Detection}

\author{Sandro Papais$^{1}$\thanks{Work done an intern at Zoox.\\ Correspondence: sandro.papais@robotics.utias.utoronto.ca } \qquad Lezhou Feng$^2$ \qquad Charles Cossette$^2$ \qquad Lingting Ge$^2$ \\
$^1$University of Toronto \qquad $^2$ Zoox Inc.
}

\begin{document}
\maketitle
\begin{abstract}
Vision Transformers (ViTs) enable strong multi-view 3D detection but are limited by high inference latency from dense token and query processing across multiple views and large 3D regions. Existing sparsity methods, designed mainly for 2D vision, prune or merge image tokens but do not extend to full-model sparsity or address 3D object queries. We introduce \myacro, a relevance-aligned sparsity framework that jointly selects 2D image tokens and 3D object queries while storing filtered features for reactivation. Mutual 2D–3D relevance heads allocate compute to driving-critical content and preserve other embeddings. Evaluated on nuScenes and our new nuScenes-Relevance benchmark, \myacro\ achieves up to 3$\times$ faster inference with marginal accuracy loss, establishing real-time large-scale ViT-based 3D detection while maintaining accuracy on planning-critical agents.
\end{abstract}

\section{Introduction}
\label{sec:intro}

Transformers dominate modern perception, yet dense attention over long image sequences and large 3D search spaces remains a barrier to real-time deployment. In autonomous driving, where latency and safety are critical, the challenge is not only to \emph{reduce} computation but to \emph{allocate} it selectively. Vision Transformer (ViT)~\citep{dosovitskiy2020image} backbones and Detection Transformer (DETR)~\citep{carion2020end} decoders achieve strong 3D perception but incur quadratic costs over tokens and queries, even though urban scenes are dominated by background (sky, road, buildings) and agents that are inconsequential for motion planning. Uniform computation is thus wasteful, treating all tokens and candidate objects as equally important and misaligning perception with the downstream prediction and planning~\citep{wang2025trendsmotionpredictiondeployable}.

Prior efficiency work focuses on one modality in isolation. ViT sparsity methods prune or merge \emph{image tokens} for 2D tasks~\citep{rao2021dynamicvit,bolya2023tokenmerging,liu2024revisiting,huang2025tokencropr}, while DETR variants suppress \emph{encoder tokens} or \emph{decoder queries} for 2D detection~\citep{roh2021sparse,zheng2023less}. For multi-view 3D detection, ToC3D~\citep{zhang2024make} compresses tokens using historical queries. None of these methods provides end-to-end sparsity over both 2D tokens and 3D queries or accounts for planner relevance, and benchmarks such as nuScenes~\cite{caesar2020nuscenes} weight all annotated agents equally, so errors on distant or non-interacting actors can dominate the metrics, and there is no way to focus evaluation on the most relevant agents.

We introduce \myacro, a \emph{planner-aligned} sparsity framework that scores and routes both image tokens and 3D object queries with lightweight 2D--3D relevance heads. Existing transformer sparsity methods act only on \emph{image tokens} in 2D ViT backbones or on \emph{object queries} in DETR-style decoders, and do not jointly sparsify 2D tokens and 3D queries in a planner-aligned way. In contrast, \myacro\ learns a unified relevance function supervised by a planner-inspired future interaction corridor, routes high-relevance queries to deeper layers, and stores lower-relevance queries in feature buffers for selective reactivation. Relative to token compression approaches~\cite{zhang2024make}, this store--reactivate design avoids merge--unmerge overhead, works on the \emph{first frame}, reduces both $\mathcal{O}(N^2)$ backbone and decoder attention, and maintains accuracy under more aggressive sparsity budgets than previous methods.

Our contributions are: \textbf{(i)} \textbf{Unified end-to-end sparsity} that \emph{jointly} prunes tokens and queries within a single architecture. \textbf{(ii)} \textbf{Planning-aligned relevance} supervised by \emph{future interaction corridors} capturing short-horizon ego--agent proximity, aligning perception budgets with planning. \textbf{(iii)} \textbf{Real-time ViTs at scale} via store--reactivate buffers that preserve recoverability under aggressive sparsity, enabling further latency reduction. \textbf{(iv)} \textbf{nuScenes-Relevance} (nuScenes-R), a relevance-aware evaluation protocol on nuScenes that measures accuracy specifically on planning-critical agents defined by future interaction corridors. \myacro\ reduces latency by up to \(3{\times}\) with marginal accuracy loss, enabling real-time ViT-based multi-view 3D detection.
\begin{figure*}[t]
    \centering
    \includegraphics[width=0.9\linewidth]{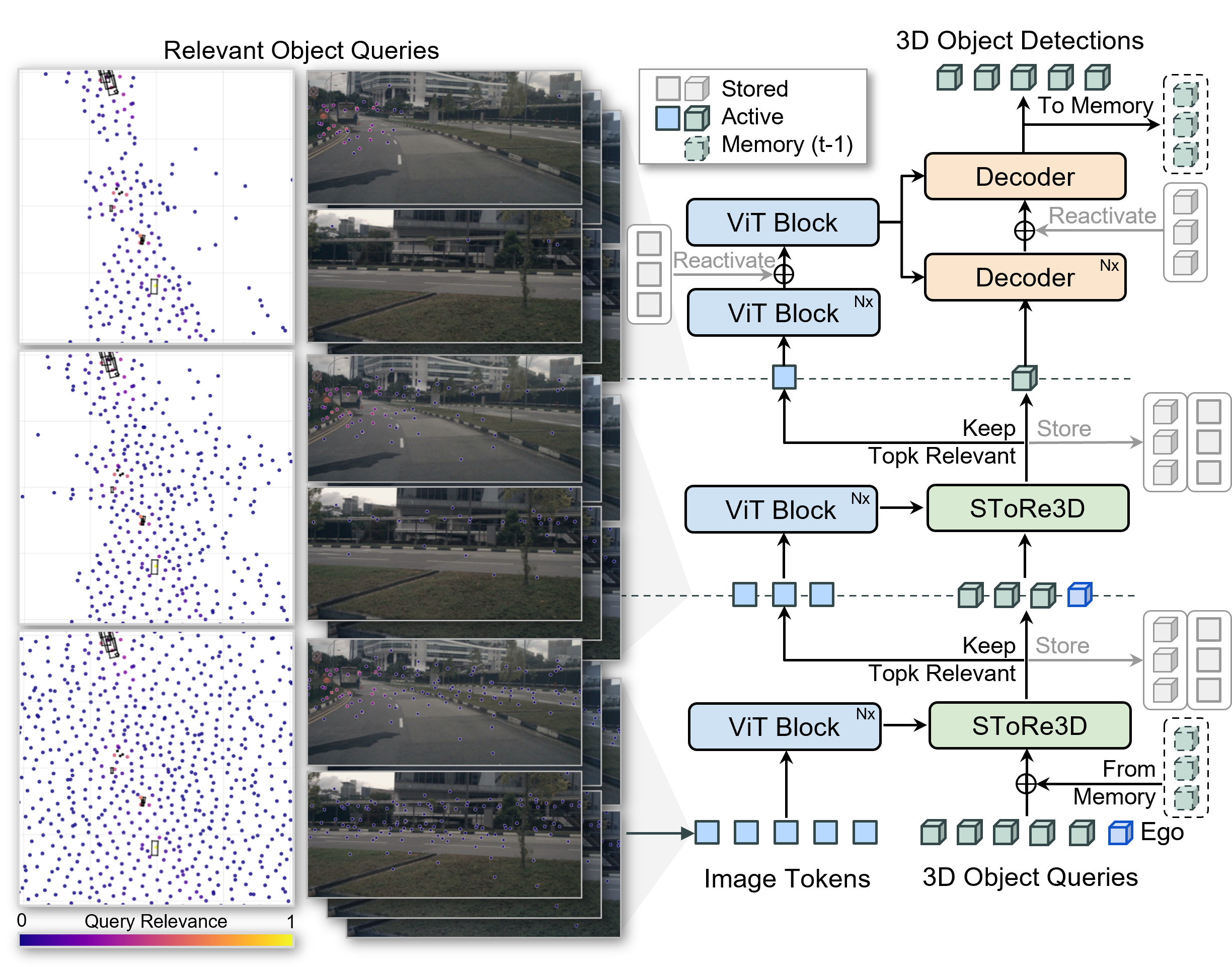}
    \caption{\myacro\ routes computation via relevance: tokens and queries above stage-wise thresholds are processed further, while the rest are \emph{stored} for reactivation. The BEV plots and camera images show how queries are reduced across stages based on their relevance scores.}
    \label{fig:intro}
\end{figure*}


\section{Related Work}

\vspace{-0.5em}
\paragraph{Efficient multi-view 3D object detection}
Early camera-only 3D detectors lifted multi-view image features into BEV space before aggregation~\citep{philion2020lift}. Transformer-based methods such as DETR3D~\citep{wang2022detr3d}, and PETR~\citep{liu2022petr} introduced 3D queries to attend across views, while BEVFormer~\citep{li2022bevformer}, Sparse4D~\citep{lin2023sparse4d}, SparseBEV~\citep{liu2023sparsebev}, PointBEV~\citep{chambon2024pointbev}, and StreamPETR~\citep{wang2023exploring} further improved efficiency and performance through sparse feature sampling, BEV representations, deformable attention, and temporal aggregation. However, real-time deployment with large ViTs remains challenging. Prior works explore sparsity either in ViT backbones or in DETR-style encoder/decoders, but treat these axes largely in isolation. As summarized in Table~\ref{tab:rw}, \myacro\ is, to our knowledge, the first to provide end-to-end query and key sparsity across both the ViT backbone and DETR3D decoder for multi-view 3D detection.

\paragraph{ViT token sparsity} 
Transformer efficiency has been pursued via approximate attention~\citep{choromanski2020rethinking,wang2020linformer,dao2022flashattention,ainslie2023gqa}, component pruning~\citep{voita2019analyzing,michel2019sixteen,meng2022adavit}, and vision-specific inductive biases~\citep{mehta2021mobilevit,graham2021levit,liu2021swin}. For ViTs, token \emph{pruning} approaches learn saliency to drop patches progressively~\citep{rao2021dynamicvit,liang2022not,xu2022evovit,fayyaz2022adaptive,yao2022spvit,li2022savit,yu2024synergistic,xu2023xpruner}, \emph{learned tokenization} approaches select informative latent tokens~\citep{ryoo2021tokenlearner,tang2022patch}, and \emph{merging/fusion} approaches reduce redundancy by grouping similar tokens~\citep{bolya2023tokenmerging,xu2024tokenfusion,lee2024learning,lee2024multi}. Extensions to dense tasks such as detection exist~\citep{liu2024revisiting,huang2025tokencropr}. However, these methods operate on \emph{image tokens only}, assume 2D salience, and provide no mechanism to coordinate with 3D object queries, which is essential for multi-view 3D detection.

\paragraph{DETR token sparsity} 
Efficiency in DETR-style detectors is typically achieved by sparsifying encoder tokens or decoder queries. Deformable DETR replaces global attention with sparse, reference-point sampling~\citep{zhu2020deformable}. Sparse-DETR~\citep{roh2021sparse}, Focus-DETR~\cite{zheng2023less}, IMFA~\cite{zhang2023towards}, and Salience-DETR~\cite{hou2024salience} further limit token or query processing through learned salience. DN-/DAB-/DINO-/RT-DETR variants~\citep{li2022dn,liu2022dab,zhang2022dino,zhao2024detrs,yao2021efficient} primarily accelerate convergence via query initialization and denoising rather than structural sparsity. For multi-view 3D detection, FocalPETR selects foreground tokens for the decoder with a 2D auxiliary head~\citep{wang2023focal}, while ToC3D compresses backbone tokens using history-driven scores and merge--unmerge routing~\citep{zhang2024make}. These approaches remain largely \emph{token-only}, focusing on the backbone or encoder while leaving decoder query redundancy under-exploited. ToC3D’s reliance on temporal priors also limits first-frame efficiency, and its per-block regrouping introduces additional overhead. In contrast, \myacro\ applies \emph{joint} 2D--3D sparsity, works from the first frame, and avoids merge--unmerge complexity via lightweight \emph{store--reactivate} buffers. 
\begin{table}[!htb]
\centering
\resizebox{1\linewidth}{!}{
\begin{tabular}{ll|cc|cc}
\toprule 
\multicolumn{2}{c|}{Visual Sparsity Approaches} & \multicolumn{2}{c|}{ViT}  & \multicolumn{2}{c}{DETR} \\
2D Detection        & 3D Detection  & QS & KS & QS & KS \\ \midrule
SwinT\cite{liu2021swin}, ViTDet~\cite{li2022exploring}        & --            &       & \checkmark &        &        \\
SViT~\cite{liu2024revisiting}, Cropr~\cite{huang2025tokencropr}          & ToC3D~\cite{zhang2024make}        & \checkmark & \checkmark &        &        \\
DeformableDETR~\cite{zhu2020deformable}      & BEVFormer~\cite{li2022bevformer}    &       &       &        & \checkmark \\
SparseDETR~\cite{roh2021sparse} & FocalPETR~\cite{wang2023focal} &       &       & \checkmark & \checkmark \\ \midrule
\rowcolor{highlight} --                    & \myacro  & \checkmark & \checkmark & \checkmark & \checkmark \\ \bottomrule
\end{tabular}
}
\caption{Comparison of visual sparsity approaches. QS: query sparsity. KS: key sparsity. In ViTs, QS and KS operate on image tokens. In DETR decoders, QS sparsifies object queries and KS samples encoder image tokens. \myacro\ jointly sparsifies queries and keys in both the ViT backbone and DETR decoder.}
\label{tab:rw}
\end{table}

\vspace{-1em}
\paragraph{Planning- and safety-critical perception}
Beyond efficiency, recent works align perception with downstream decision-making by emphasizing agents that matter most, including risk-object identification~\citep{li2020make,li2023droid}, planner-guided attention~\citep{wei2021perceive}, and end-to-end perception--prediction--planing frameworks such as ForeSight~\citep{papais2025foresight}, UniAD~\citep{hu2023planning}, and SparseDrive~\cite{sun2025sparsedrive}. These methods couple perception to planning objectives but lack a scalable \emph{architectural} mechanism for token and query sparsity with ViTs. \myacro\ closes this gap by supervising sparsity with a \emph{future interaction corridor} and evaluating on nuScenes-R.

\vspace{-0.5em}
\section{Method}
\vspace{-0.5em}
\label{sec:method}

\myacro\ applies \emph{joint, hierarchical sparsity} to both image tokens and 3D object queries in a temporal multi-view 3D detector~\cite{wang2023exploring}. At each stage, lightweight relevance heads assign each image token and 3D object query a scalar \emph{relevance} score. We consider two ways to define and supervise this relevance: a \emph{planning-aligned} variant $r^{\text{plan}}$, which focuses on objects the ego vehicle may need to react to in the near future (e.g., lead vehicles, crossing pedestrians), and a \emph{detection-aligned} variant $r^{\text{det}}$, which aims to keep all foreground objects while rejecting background clutter. The most relevant tokens and queries are kept active and propagated to deeper layers, while low-relevance embeddings are written to lightweight \emph{storage buffers} instead of being discarded. Buffered embeddings are later \emph{reactivated}, yielding a \emph{store--reactivate} form of sparsity that avoids irreversible pruning.
\begin{figure*}[!ht]
    \centering
    \includegraphics[width=0.95\linewidth]{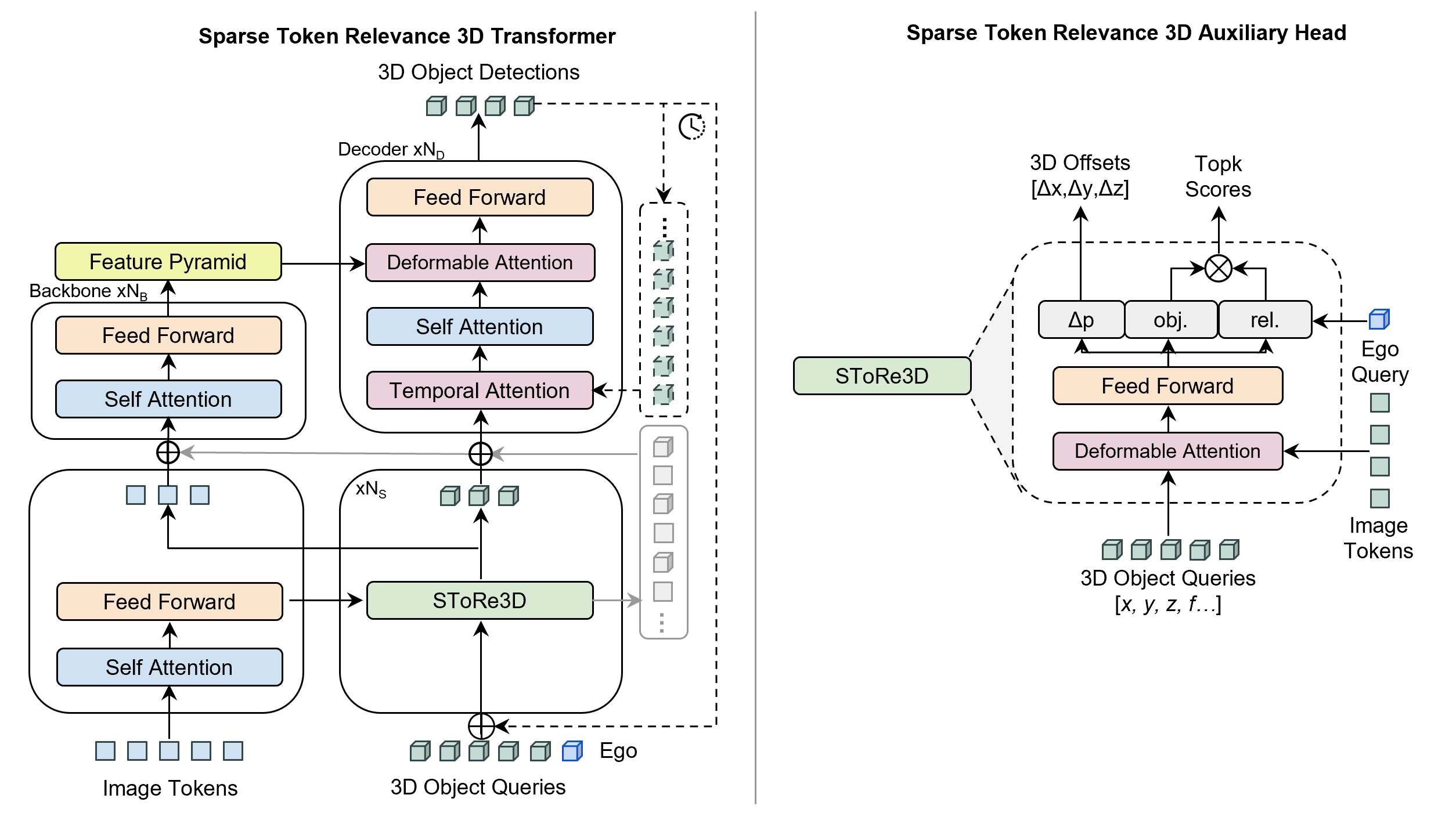}
    \caption{\myacro\ architecture builds on a ViT backbone and DETR3D-style decoder with relevance heads that score image tokens and 3D queries at each stage, enabling joint sparsity across backbone and decoder.}
    \label{fig:arch}
\end{figure*}

\subsection{Problem Formulation}
\label{sec:method1}

We consider multi-view 3D detection with $V$ synchronized cameras over a temporal window $\{t\!-\!T,\dots,t\}$. Each view produces tokens $\mathbf{X}_{t,v}$ from a ViT backbone, concatenated as $\mathbf{X}_{t}$. The backbone interleaves global and windowed attention, and a feature pyramid network (FPN) provides multi-scale features. Detection object queries $\mathbf{Q}_{t}$ are anchored at 3D positions $\mathbf{p}=(x,y,z)$, initialized as $\mathbf{q}^{(0)}=\mathrm{MLP}(\mathrm{PE}(\mathbf{p}))$, and refined by a multi-scale deformable DETR-style decoder that applies deformable cross-attention over the FPN features.

Following streaming detection and tracking works~\citep{wang2023exploring,lin2023sparse4d, papais2024swtrack, cheong2026scatr}, top-$K$ queries propagate across frames, maintained in a temporal memory with temporal reference points transformed to the current ego frame. The active query set combines propagated and initialized queries, $N_d = N_{\mathrm{prop}} + N_{\mathrm{init}}$. For the first frame, additional initialized queries replace the propagated queries. This yields unified token $\mathbf{X}_t$ and query $\mathbf{Q}_t$ sets, which serve as input to the sparse relevance module.

\subsection{Defining Object Relevance}
\label{sec:method2}

Intuitively, we call an agent \emph{planning-critical} if the ego vehicle may need to react to it in the near future, for example, if the agent will pass close to any plausible ego trajectory over the next few seconds. We formalize this by defining a \emph{future interaction corridor} in BEV around the ego vehicle's candidate motion and labeling agents that enter this corridor as relevant. Let $\mathcal{B}_{\mathrm{ego}}(\tau)$ and $\mathcal{B}_i(\tau)$ denote oriented boxes for ego and agent-$i$ at $t+\tau$. Swept sets over a horizon $H$ are defined by the convex hull of the union of future boxes:
\begin{equation}
\mathcal{S}_{i}(H) = \mathrm{conv}\Big( \bigcup_{\tau\in[0,H]}\mathcal{B}_{i}(\tau)\Big).
\label{eq:swept}
\end{equation}
An agent is \emph{relevant} if the closest distance between its swept polygon and the ego swept polygon is within a margin $d_{\min}$:
\begin{equation}
y_i^{\mathrm{rel}} = \mathbbm{1}\!\Big(\mathrm{dist}\!\big(\mathcal{S}_{i}(H),\mathcal{S}_{\mathrm{ego}}(H)\big) \le d_{\min}\Big).
\label{eq:relevance}
\end{equation}
This polygonal corridor captures translation and orientation over discrete \emph{future} steps for $H{=}5$\ seconds, and the labels $\{y_i^{\mathrm{rel}}\}$ supervise relevance. The same definitions apply to nuScenes-R metrics (Section~\ref{sec:benchmark}).

\subsection{Unified 2D-3D Relevance Prediction}
\label{sec:method3}

We predict planning-aligned relevance for both modalities using \emph{mutual gating}: queries are scored in the context of tokens and vice versa. Query relevance is supervised by corridor labels, while token relevance is aggregated from query attention. For object query $\mathbf{q}_j$, we compute a context vector from deformable cross-attended tokens and optionally an ego embedding $\mathbf{e}_t$,
\begin{equation}
\mathbf{c}^{\mathrm{qry}}_j = \mathrm{CrossAttn}_{\text{def}}(\mathbf{q}_j,\mathbf{X}_t)\oplus \mathbf{e}_t,
\end{equation}
\begin{equation}    
r^{\mathrm{qry}}_j = \sigma\!\Big(\mathbf{u}^\top \phi([\mathbf{q}_j \| \mathbf{c}^{\mathrm{qry}}_j])\Big).
\end{equation}
where $\phi$ is a small MLP, $\oplus$ indicates optional concatenation for planning relevance, $\sigma$ is the sigmoid function, and $\mathrm{CrossAttn}_{\text{def}}$ denotes the same multi-scale deformable cross-attention used in the detection decoder, attending to a sparse set of FPN features around the query reference points. The ego term lets $r^{\mathrm{qry}}$ condition relevance on ego--agent motion. Image token relevance $r^{\mathrm{img}}_i$ aggregates attention from top-$K$ relevant queries:
\begin{equation}
r^{\mathrm{img}}_i = \tfrac{1}{K}\!\sum_{j\in \mathcal{K}^{\mathrm{qry}}}\! A_{j\rightarrow i},
\end{equation}
yielding a query-aware token relevance that emphasizes regions supported by high-relevance 3D queries. The scores $r^{\mathrm{qry}}_j$ and $r^{\mathrm{img}}_i$ serve as routing signals for stage-wise sparsification (Section~\ref{sec:method4}). We supervise $r^{\mathrm{qry}}$ with binary labels $y^{\mathrm{rel}}$ from the interaction corridors (Section~\ref{sec:method2}).

\subsection{Hierarchical Token Storage}
\label{sec:method4}

Conceptually, we split tokens and queries into two groups at each stage: a small active set that continues through the network, and a temporarily inactive set that is cached in a buffer rather than discarded. Joint sparsity is applied to both the \emph{backbone token stream} and the \emph{query stream} within the backbone and encoder. After each stage $\ell$, we \emph{filter} tokens/queries using the relevance scores from Section~\ref{sec:method3}, \emph{store} the remainder in buffers, and later \emph{reintroduce} them at the final layer. For stage-wise filtering and storage, let $N_\ell$ and $Q_\ell$ be the numbers of tokens and queries at stage $\ell$. We keep fractions $\rho^{\mathrm{img}}_\ell,\rho^{\mathrm{qry}}_\ell\in(0,1]$ via Gumbel-softmax top-$k$~\citep{jang2016categorical}, and the filtered items are written to buffers:
\begin{equation}
\mathcal{K}_\ell = \mathrm{TopK}(\mathbf{r}_{\ell},\lfloor \rho_\ell N_\ell\rfloor).
\end{equation}
The fractions $\rho^{\mathrm{img}}_\ell,\rho^{\mathrm{qry}}_\ell\in(0,1]$ follow a non-increasing \emph{hierarchical schedule} with depth and are regularized toward targets in Section~\ref{sec:method5}. Let $\overline{\mathcal{K}}^{\mathrm{img}}_\ell$ and $\overline{\mathcal{K}}^{\mathrm{qry}}_\ell$ be complements of the kept indices. We write filtered features to buffers immediately after filtering and before the next stage:
\begin{equation}
\mathbf{S}^{\mathrm{img}}_{\ell}\!\leftarrow\!\mathbf{X}_\ell[\overline{\mathcal{K}}^{\mathrm{img}}_\ell],\qquad
\mathbf{S}^{\mathrm{qry}}_{\ell}\!\leftarrow\!\mathbf{Q}_\ell[\overline{\mathcal{K}}^{\mathrm{qry}}_\ell].
\end{equation}
Training under aggressive sparsity is non-trivial: as the number of active tokens and queries shrinks, the model receives fewer supervised examples per layer, making it harder to learn stable relevance scores and high-quality features. Naively pruning hard from the beginning often leads to collapsed solutions or severe underfitting. To mitigate information loss, we \emph{reactivate} stored image tokens and object queries for the final attention layer of the backbone and decoder using the updated context. We use a two-level storage schedule: (i) \emph{depth-wise} budgets $(\rho^{\mathrm{img}}_\ell,\rho^{\mathrm{qry}}_\ell)$ are non-increasing with $\ell$; (ii) \emph{training-time} pruning is introduced gradually (linear schedule from dense to target budgets) to maintain stability at high sparsity without finetuning. This preserves global context without restoring all pruned items and adds negligible cost.

\subsection{Optimization Approach}
\label{sec:method5}

The overall framework is trained end-to-end with a multi-task objective over detection, relevance, and auxiliary ROI supervision. For detection, we use a combination of focal loss~\cite{lin2017focal} for classification and L1 loss for bounding box regression with Hungarian bipartite matching. Query relevance is trained with a Gaussian focal loss~\cite{law2018cornernet} on predicted scores $r^{\text{qry}}$ and binary relevance labels $y^{\mathrm{rel}}$. For the planning-aligned variant, we supervise $r^{\text{plan}}$ using labels derived from the future interaction corridor defined in Section~\ref{sec:method2}: agents that enter the corridor are labeled relevant, all others irrelevant. For the detection-aligned variant, we instead supervise $r^{\text{det}}$ using standard foreground/background labels from the 3D detection head (e.g., treating queries matched to ground-truth boxes as relevant). To prevent the relevance heads from interfering with feature learning in the main detector, we stop gradients from the relevance heads at the input object queries and image tokens. Therefore, end-to-end gradients to tokens and queries flow only through the detection loss. In addition, an auxiliary loss is used to supervise ROI feature extraction, with classification and regression terms on 2D image-space targets~\cite{wang2023focal}. The joint loss is
\begin{equation}
\mathcal{L} = \mathcal{L}_{\text{det}} +
\lambda_{\text{rel}}\mathcal{L}^{\text{qry}}_{\text{rel}} +
\lambda_{\text{aux}}\mathcal{L}_{\text{aux}}
\end{equation}
where $\lambda_{\text{rel}}$ and $\lambda_{\text{aux}}$ are balancing weights and $\mathcal{L}_{\text{det}}$ includes focal and L1 losses with Hungarian matching. Token relevance is indirectly supervised through cross-attention with the queries. Gumbel-TopK provides differentiable routing, with the pruning budget linearly increased over training iterations from dense (no sparsity) to the target sparsity.

\vspace{-0.5em}
\section{Benchmarking Relevance}
\vspace{-0.5em}
\label{sec:benchmark}

\begin{figure}[!ht]
    \centering
    \includegraphics[width=0.95\linewidth]{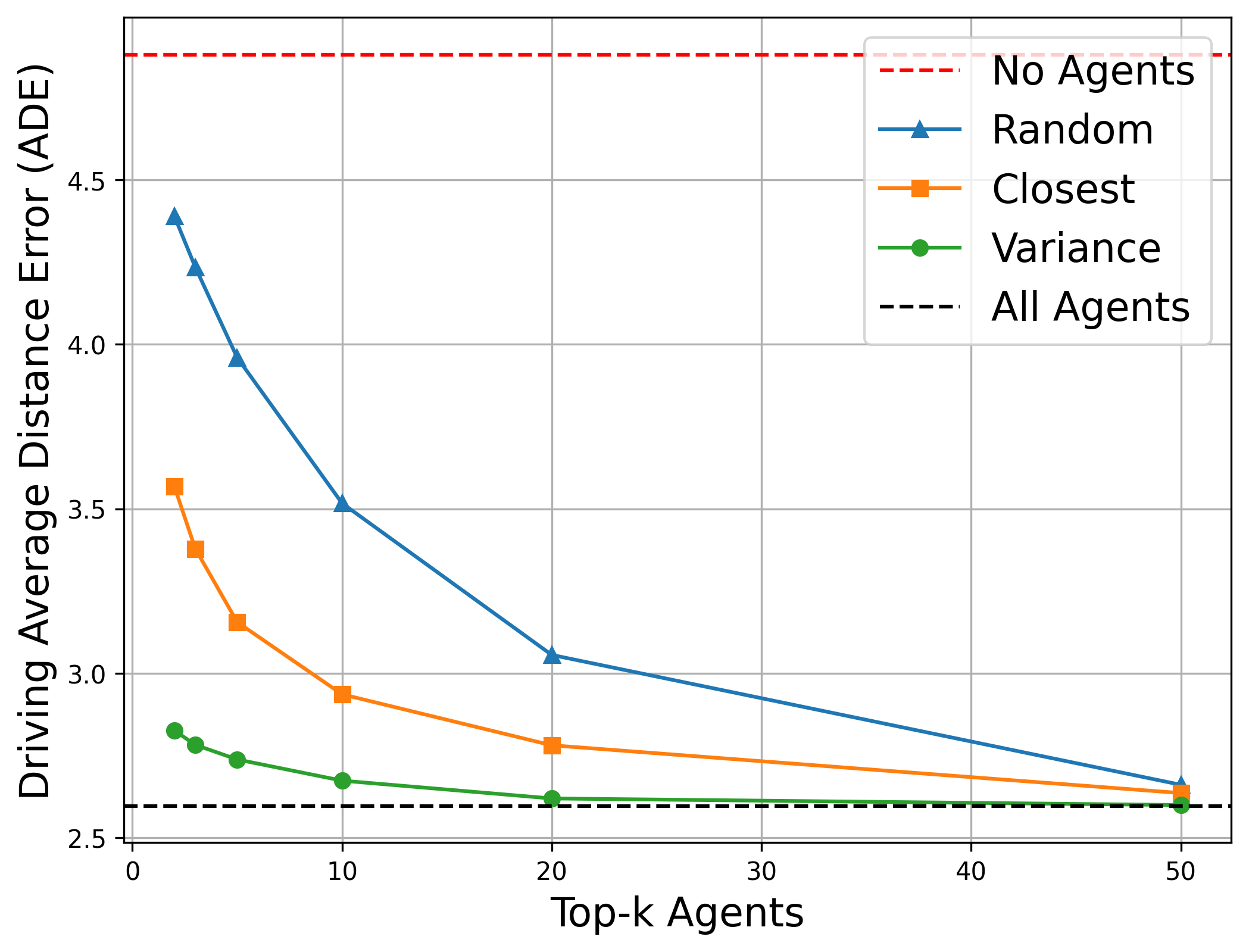}
    \caption{Planner performance vs.\ retained agents using a fixed pretrained motion-planning network for analysis only. Planning performance saturates with a subset of agents, motivating relevance-based compute allocation.}
    \label{fig:plan_analysis}
\end{figure}

\paragraph{Why relevance}
Conventional detectors expend equal compute on all agents, inflating latency and misaligning perception with planning, even though many urban objects (e.g., parked vehicles, distant pedestrians) are inconsequential for near-term driving. We therefore evaluate perception under a \emph{relevance-driven} lens: prioritize agents that matter for planning. To quantify this, we vary the number of detected agents provided to a fixed pretrained motion-planning network used only for analysis, not during \myacro\ training or inference. Performance saturates with only $10$–$20$ agents (Figure~\ref{fig:plan_analysis}), indicating substantial headroom to reduce compute without harming planning.

\begin{figure}[!ht]
    \centering
    \includegraphics[width=0.94\linewidth]{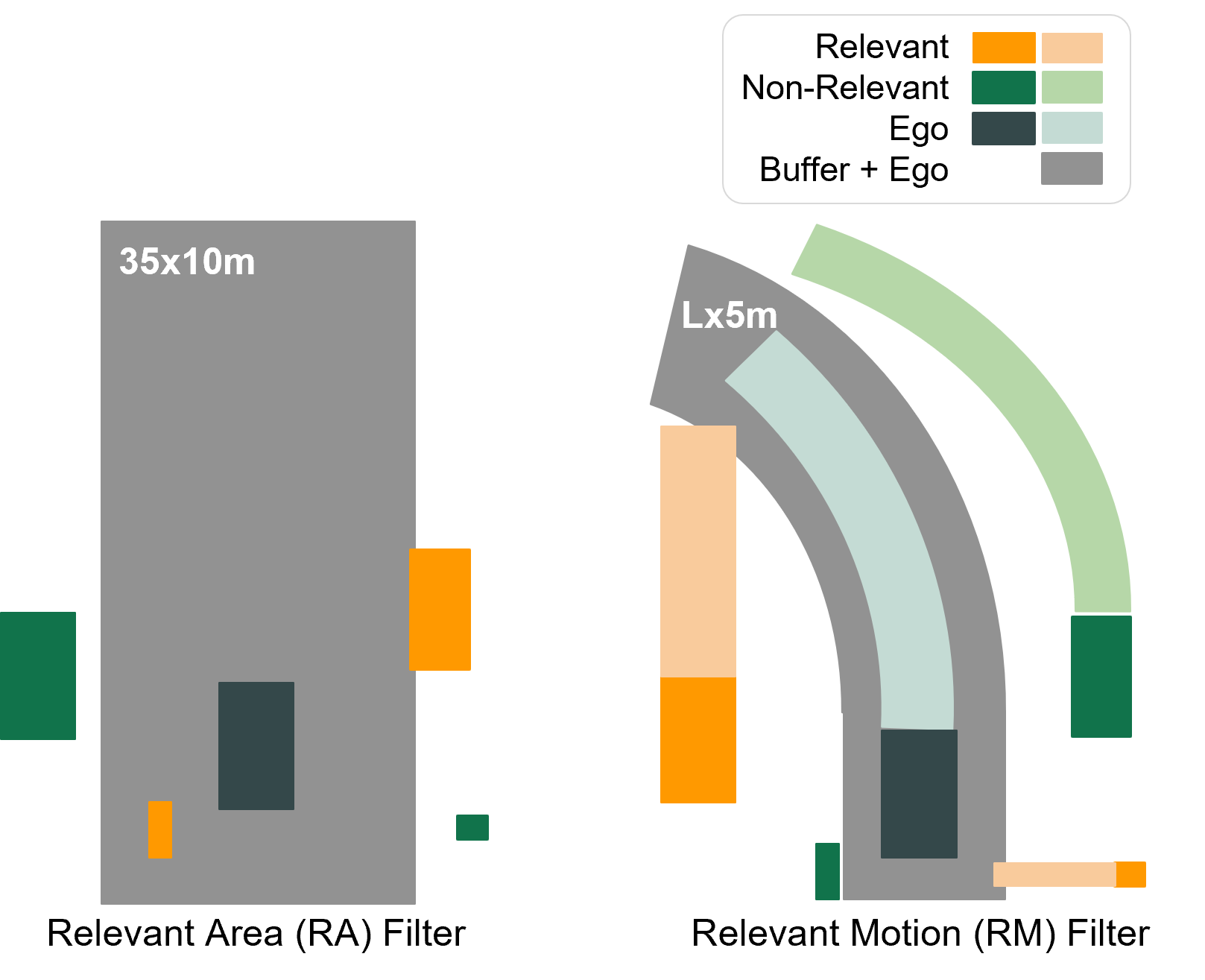}
    \caption{Relevant object labeling example geometry.}
    \label{fig:rel}
\end{figure}

\paragraph{Planning-relevant labels}
Ground-truth relevance follows the future interaction corridors defined in Section~\ref{sec:method2}. For ego and agent-$i$, we construct swept sets $\mathcal{S}_{\mathrm{ego}}(H)$ and $\mathcal{S}_i(H)$ over a horizon $H{=}5$ seconds and label an agent relevant if the closest distance between the buffered corridors is below a margin $d_{\min}$. We fix a single operating point by choosing $d_{\min}$ as the 10th percentile of ego–agent distances on nuScenes, yielding $d_{\min}{\approx}1.2$ m, $\sim$3 relevant agents per frame on average, and at most $\sim$30. Labels are generated using ground truth trajectories, convexifying swept polygons, and applying the buffered-intersection test (Figure~\ref{fig:rel}).
\vspace{-0.5em}

\paragraph{Relevance metrics}
Standard detection metrics, mean average precision (mAP) and nuScenes detection score (NDS), treat all agents equally, regardless of planning importance. We instead define relevance via a \emph{future interaction corridor}: 5-second swept polygons for ego and each agent. An agent is labeled relevant if the closest distance $d_C$ between its corridor and the ego’s corridor is below a buffer $d_{RM}$. Empirically, $d_{RM}{=}1.2$ m selects $\sim$10\% of agents, or about 3 relevant objects per frame on average. This definition underlies our benchmark, nuScenes-R.
\begin{figure*}[!ht]
    \centering
    \includegraphics[width=0.8\linewidth]{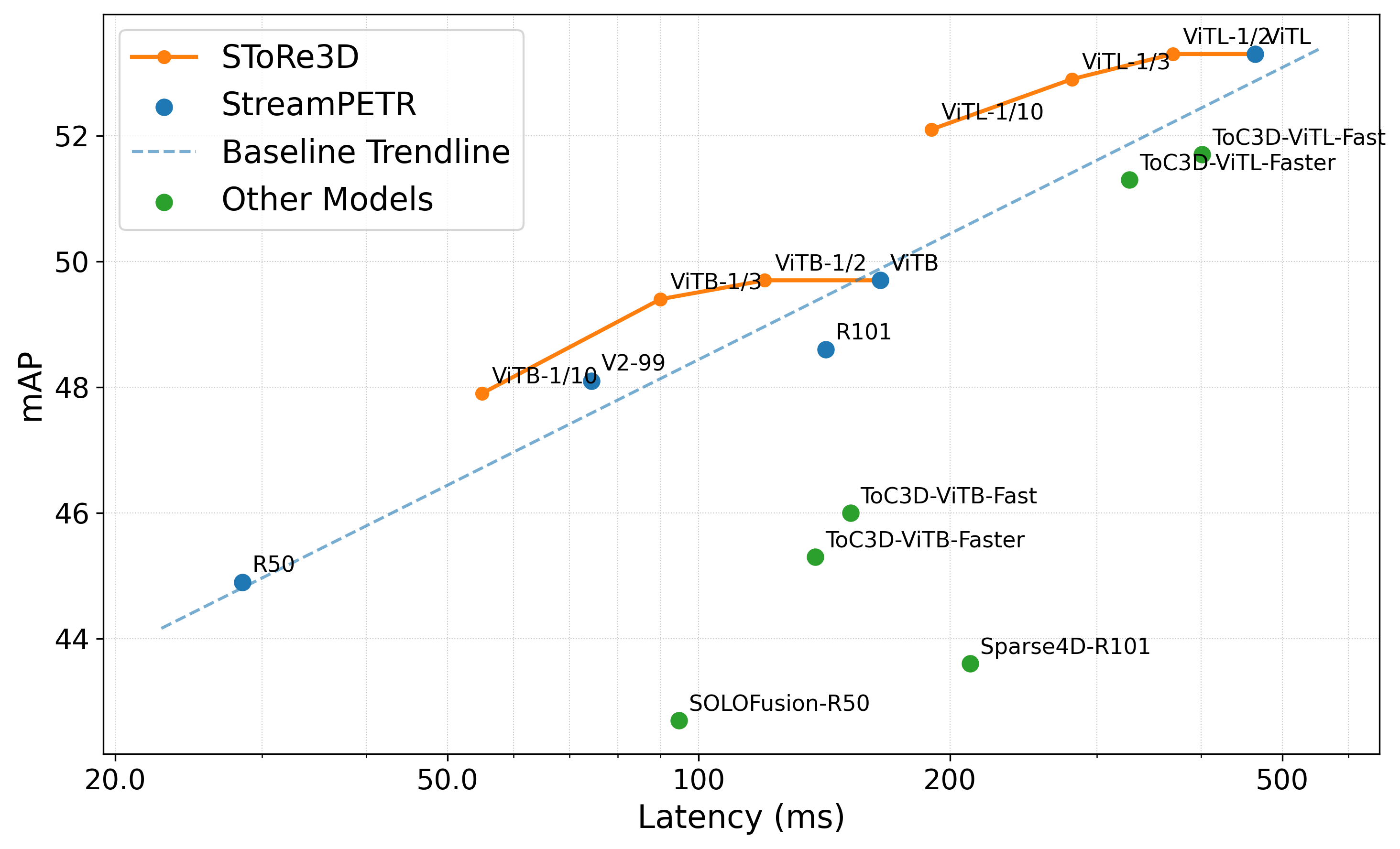}
    \caption{Latency-accuracy curves for \myacro\ under varying sparsity, compared to alternatives. Each point corresponds to a different keep ratio. Increasing sparsity moves models toward higher FPS with only modest drops in mAP.}
    \label{fig:mAP}
\end{figure*}

We report two variants: \emph{relevant motion} (RM) filtered metrics (NDS-RM), which apply the RM filter as described above, and \emph{relevant area} filtered metrics (mAP-RA, NDS-RA), which use a fixed detection area around the vehicle for evaluation. Because RM relies on privileged future information, detections are matched to RM-filtered ground truth for true positives and false positives, but false negatives are difficult to compute. Together, RM and RA ensure that \myacro’s relevance-adaptive sparsity is evaluated fairly, measuring whether accuracy is preserved on \emph{planning-critical} agents while enabling substantial efficiency gains.

\vspace{-0.5em}
\section{Experiments}

\subsection{Experiment Setup}
\vspace{-0.5em}

\paragraph{Dataset} 
We evaluate on the nuScenes 3D detection benchmark~\citep{caesar2020nuscenes}, which contains 1,000 $\sim$20s scenes at 20 Hz with six surround cameras per sample, calibrated with known intrinsics and extrinsics. Annotations are provided every 0.5 s, yielding 28k/6k/6k samples for train/val/test over ten classes. We report standard nuScenes metrics (mAP, NDS) and relevance-filtered metrics on planning-critical agents using nuScenes-R (Section~\ref{sec:benchmark}), including Relevant-Area (RA) and Relevant-Motion (RM) variants.

\paragraph{Implementation Details}
We use six synchronized cameras with standard calibration. Unless noted otherwise, \myacro\ is implemented and evaluated with ViT-B and ViT-L backbones~\citep{dosovitskiy2020image} initialized from EVA-02~\citep{fang2024eva}. ResNet-50/101~\citep{he2016deep} and V2-99~\citep{lee2019energy} appear only as reproduced baselines reported from prior work. Unless stated otherwise, input resolution is \(320\times 800\); we also report \(256\times 704\), \(512\times 1408\), and \(800\times 1600\) for accuracy–speed trade-offs. The detector follows a DETR-style design with multi-scale features, embedding dimension $D{=}256$, and $L{=}6$ decoder layers. The baseline uses 644 detection queries and 256 temporal queries (900 total) with four frames of memory. Denoising~\citep{wang2023focal,wang2023exploring} is applied during training. Models are trained for 24 epochs on 8$\times$A100 GPUs (batch size 16); inference latency is measured at batch size 1 on a single RTX3090. We use AdamW with cosine decay and mixed precision. Relevance heads are two-layer MLPs with GELU and $\mathrm{TopK}$ gating that use differentiable Gumbel-softmax. We report three operating points, \myacro-1/2, \myacro-1/3, and \myacro-1/10, corresponding to hierarchical schedules that retain roughly half, one-third, and one-tenth of tokens/queries.
\begin{table*}[!htb]
\centering
\begin{tabular}{@{}l|c|cc|ccc|c@{}}
\toprule
\textbf{Methods}    & \textbf{Backbone$\times$W-S} & \textbf{mAP$\uparrow$} & \textbf{-RA$\uparrow$} & \textbf{NDS$\uparrow$} &  \textbf{-RA$\uparrow$} & \textbf{-RM$\uparrow$} & \textbf{FPS$\uparrow$} \\ \midrule
StreamPETR                                                        & V2-99$\times$800      & 0.482         & 0.605         & 0.571         & 0.646         & 0.428         & 13.5          \\ \midrule \midrule
BEVFormer                                                         & R101-DCN$\times$1600  & 0.416         & -             & 0.517         & -             & -             & 3.3           \\
Sparse4D                                                          & R101-DCN$\times$1600  & 0.444         & -             & 0.550         & -             & -             & 4.7           \\ \midrule
SOLOFusion                                                        & R101$\times$1408      & 0.483         & -             & 0.582         & -             & -             & -           \\ 
StreamPETR                                                        & R101$\times$1408      & 0.486         & -             & 0.578         & -             & -             & 7.0           \\ 
SparseBEV                                                        & R101$\times$1408      & 0.501         & -             & 0.592         & -             & -             & -           \\ 
Sparse4Dv2                                                        & R101$\times$1408      & 0.505         & -             & 0.598         & -             & -             & -           \\ \midrule \midrule
StreamPETR                                                        & ViT-B$\times$800      & 0.497         & 0.627         & 0.584         & 0.667         & 0.443         & 6.1           \\ \midrule
ToC3D-Fast                                                        & ViT-B$\times$800-1/2  & 0.46          & 0.615        & 0.562         & 0.664         & 0.431         & 6.6           \\
\rowcolor{highlight} \myacro-1/2  & ViT-B$\times$800-1/2  & 0.493         & 0.627         & 0.581         & 0.665         & 0.441         & 8.2           \\ \midrule
ToC3D-Faster                                                      & ViT-B$\times$800-1/3  & 0.453         & 0.618         & 0.559         & 0.656         & 0.43          & 7.3           \\
\rowcolor{highlight} \myacro-1/3  & ViT-B$\times$800-1/3  & 0.489         & 0.623         & 0.578         & 0.65          & 0.435         & 10.6          \\ \midrule
\rowcolor{highlight} \myacro-1/10  & ViT-B$\times$800-1/10 & 0.479         & 0.612         & 0.571         & 0.639         & 0.43          & 17.7          \\ \midrule \midrule
StreamPETR                                                        & ViT-L$\times$800      & 0.521         & 0.641         & 0.608         & 0.688         & 0.485         & 2.2           \\ \midrule

ToC3D-Fast                                                        & ViT-L$\times$800-1/2  & 0.523        & 0.639        & 0.610         & 0.681         & 0.463         & 2.5           \\
\rowcolor{highlight} \myacro-1/2  & ViT-L$\times$800-1/2  & 0.533         & 0.666         & 0.618         & 0.697         & 0.475         & 2.7          \\ \midrule
ToC3D-Faster                                                      & ViT-L$\times$800-1/3  & 0.517         & 0.63         & 0.609         & 0.672         & 0.453         & 3.1           \\
\rowcolor{highlight} \myacro-1/3  & ViT-L$\times$800-1/3  & 0.523         & 0.654        & 0.609         & 0.678         & 0.469         & 3.5           \\ \midrule
\rowcolor{highlight} \myacro-1/10 & ViT-L$\times$800-1/10 & 0.521         & 0.641         & 0.607         & 0.679         & 0.478         & 5.2           \\ \bottomrule
\end{tabular}
\caption{Detection performance on nuScenes and nuScenes-R validation set against SOTA methods. Methods are grouped based on comparable latency. Backbone $\times$W and -S are the image width and sparsity level, respectively. StreamPETR~\citep{wang2023exploring}, ToC3D~\citep{zhang2024make}, 
SparseBEV~\cite{liu2023sparsebev}, Sparse4Dv2~\cite{lin2023sparse4d}, SOLOFusion~\citep{park2022time}, 
BEVFormer~\citep{li2022bevformer}, and Sparse4D~\citep{lin2023sparse4d} are as reported.}
\label{tab:val}
\end{table*}

\subsection{Main Results}
\vspace{-0.5em}

\paragraph{Accuracy–efficiency trade-offs}
Figure~\ref{fig:mAP} plots the speed-–accuracy frontier of \myacro\ across sparsity regimes, alongside StreamPETR~\citep{wang2023exploring}. Jointly pruning 2D tokens and 3D queries yields monotonic FPS gains with negligible accuracy loss at low sparsity and only small losses at higher sparsity. Notably, \myacro-1/10 enables \emph{real-time} ViT-based multi-view 3D detection, running at $\sim$18 FPS with ViT-B while remaining SOTA among methods at similar latency. Under matched ViT backbone comparisons (Table~\ref{tab:val}), \myacro\ consistently improves the accuracy--latency trade-off over dense StreamPETR and token-only ToC3D operating points. As the hierarchical keep ratio decreases, inference time drops monotonically while mAP/NDS degrade smoothly, with a clear knee at mid sparsity. Pushing beyond this regime is challenging, since extreme sparsity drastically reduces the number of supervised tokens and queries per batch, \myacro's store--reactivate buffers and gradual pruning warm-up are key to maintaining stable training in this high-sparsity setting.

\paragraph{Comparison to baselines}
We compare the detection-aligned \myacro\ variant ($r^{\text{det}}$) to prior multi-view 3D detectors using standard detection metrics, and the planning-aligned variant ($r^{\text{plan}}$) using nuScenes-R. Table~\ref{tab:val} shows that \myacro\ is competitive with strong published baselines across a range of latency regimes. The controlled matched-backbone comparisons show that, relative to StreamPETR and ToC3D under the same ViT backbone family, joint token--query sparsity yields a stronger accuracy--latency trade-off than token-only compression.

\subsection{Ablation Study}
\begin{table*}[!htb]
\centering
\begin{tabular}{@{}l|c|c|c|c@{}}
\toprule
\textbf{Sparsity Approach} & \textbf{TKR} & \textbf{NDS} $\uparrow$ & \textbf{mAP} $\uparrow$ & \textbf{FPS} $\uparrow$ \\
\midrule
StreamPETR                                                            & 1            & 0.614                                  & 0.533                                  & 2.15                                     \\ \hline
$+$ Random                                                            & 0.5          & 0.567 (-7.7\%)                           & 0.465 (-12.8\%)                          & 2.45 (1.14$\times$)                      \\
$+$ DynamicViT                                                        & 0.5          & 0.597 (-2.8\%)                           & 0.505 (-5.3\%)                           & 2.47 (1.15$\times$)                      \\
$+$ ToC3D-Fast                                                      & 0.5          & 0.61 (-0.7\%)                            & 0.523 (-1.9\%)                           & 2.43 (1.13$\times$)                      \\
\rowcolor{highlight} $+$ \myacro-1/2  & 0.5          & 0.618 (+0.7\%)                            & 0.533 (0\%)                              & 2.70 (1.26$\times$)                      \\ \hline
$+$ Random                                                            & 0.3          & 0.485 (-21\%)                            & 0.36 (-32.5\%)                           & 2.9 (1.35$\times$)                       \\
$+$ DynamicViT                                                        & 0.3          & 0.593 (-3.4\%)                           & 0.493 (-7.5\%)                           & 2.92 (1.36$\times$)                      \\
$+$ ToC3D-Faster                                                    & 0.3          & 0.603 (-1.8\%)                           & 0.512 (-3.9\%)                           & 2.89 (1.34$\times$)                      \\
\rowcolor{highlight} $+$ \myacro-1/3  & 0.3          & 0.609 (-0.8\%)                           & 0.523 (-1.9\%)                           & 3.51 (1.63$\times$)                      \\ \hline
\rowcolor{highlight} $+$ \myacro-1/10 & 0.1          & 0.607 (-1.1\%)                           & 0.521 (-2.3\%)                           & 5.21 (2.42$\times$)                      \\ 
\bottomrule
\end{tabular}
\caption{Ablation of sparsity approaches \cite{wang2023exploring,rao2021dynamicvit,zhang2024make} at matched total keep ratios (TKR).}
\label{tab:abl1}
\end{table*}

On nuScenes-R, where evaluation is restricted to agents within the future interaction corridor (RM) or a fixed area around the ego (RA), \myacro\ retains strong mAP-RA/NDS-RM on planning-critical agents while running faster than dense baselines. For example, \myacro-1/10 (ViT-L) achieves 0.521 mAP, 0.607 NDS, 0.478 NDS-RM, and 5.2 FPS, compared to StreamPETR (ViT-L) at 0.521 mAP, 0.608 NDS, 0.463 NDS-RM, and 2.7 FPS. This indicates that \myacro's relevance-driven sparsity preserves accuracy where it matters most for safety-critical planning while significantly reducing latency.

\paragraph{Sparsity approach} 
Table~\ref{tab:abl1} compares alternative sparsification strategies at matched total keep ratios against our joint token–query sparsity. At \(\rho{=}0.5\) and \(\rho{=}0.3\), \myacro\ consistently achieves equal or higher accuracy at similar or lower latency than token-only approaches. The extra speedup is primarily from added query sparsity in the decoder, which ViT image token-only methods cannot realize, while tying token pruning to object queries preserves accuracy under stronger sparsity.

The store--reactivate buffers introduce modest runtime memory overhead while avoiding additional attention or recomputation. For \myacro-1/2 with ViT-B, the buffers store 64,200 pruned image/query embeddings (256-dim), adding approximately 20 MB of runtime memory while reducing compute by 397 GFLOPs. This corresponds to a 2.1\% memory overhead for a 28\% compute reduction.

\paragraph{Pruning design} 
Table~\ref{tab:abl2} ablates design choices within \myacro. Joint pruning of image tokens \emph{and} object queries (I\&O) outperforms pruning either stream alone, reducing latency in both the backbone and decoder. Retaining filtered items in \emph{store} buffers with reactivation is superior to hard \emph{pruning}, indicating that retrieval paths mitigate early pruning errors. Finally, a linear schedule (warming up from dense to target keep ratios) is more stable than flat query-reduced fine-tuning, which often fails to converge at high sparsity due to the reduced learning signal from very few active queries.
\begin{table}[!htb]
\centering
\resizebox{1\linewidth}{!}{
\begin{tabular}{@{}l|cccc|c@{}}
\toprule
\textbf{Setting} & \textbf{Train Loss} & \textbf{Pruner} & \textbf{Pruned} & \textbf{Schedule} & \textbf{mAP} $\uparrow$  \\  \midrule
None & - & - & - & - &  0.540  \\ \midrule
v1 & Top Q  & Store & I \& O & Finetune & 0.515 \\
v3 & Top Q  & Store & O & Linear & 0.534 \\
v4 & Top Q  & Store & I  & Linear & 0.527  \\
v5 & Top Q  & Remove & I \& O  & Linear  & 0.495\\ 
v6 & All Q  & Store & I \& O  & Linear & 0.513\\ \midrule
\rowcolor{highlight} \myacro & Top Q  & Store & I \& O  & Linear & 0.521 \\ \bottomrule
\end{tabular}
}
\caption{Ablation of design decisions on nuScenes validation set using ViT-L backbone for the \myacro-1/10 variant. I: image tokens. O: object queries.}
\label{tab:abl2}
\end{table}

\subsection{Qualitative Results}

\begin{figure}[!ht]
    \centering
    \includegraphics[width=0.9\linewidth]{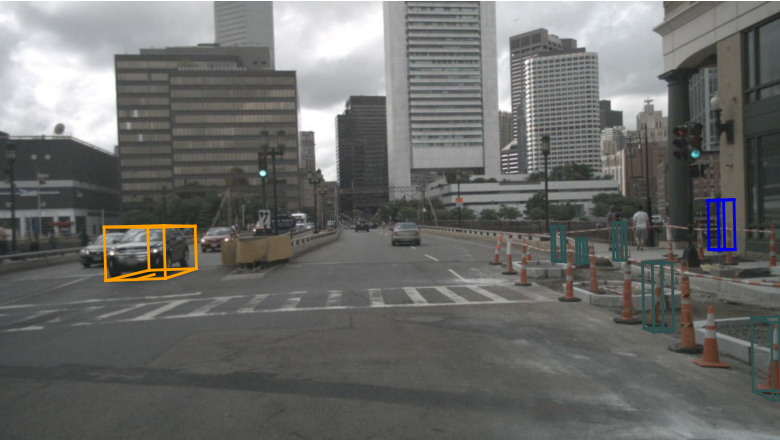}
    \includegraphics[width=0.9\linewidth]{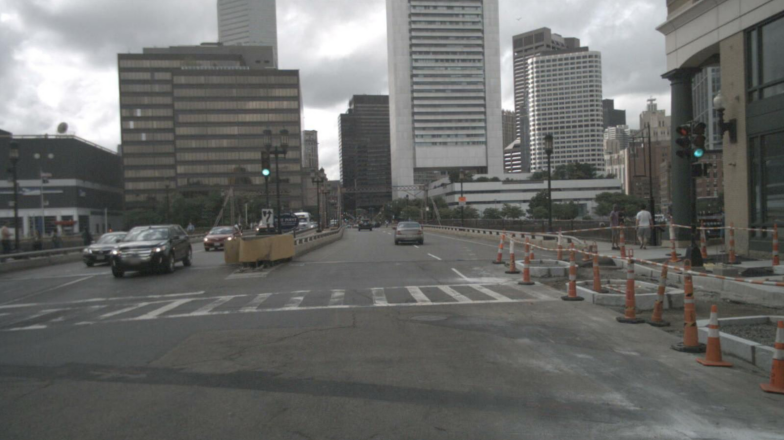}
    \caption{Visualization of false negatives for the baseline StreamPETR-R50 (top) and SToRe3D-1/10-ViT-B at similar latency (bottom).}
    \label{fig:qual}
\end{figure}

Figure~\ref{fig:qual} qualitatively compares \myacro\ to StreamPETR in challenging driving scenarios. \myacro\ suppresses background false positives by pruning low-relevance tokens and preserves detections for agents inside the interaction corridor, reducing critical false negatives and yielding more focused detections on planning-critical agents. Additional qualitative results are provided in the supplementary material.

\section{Conclusion}

We introduced \myacro, a planner-aligned sparsity framework for multi-view 3D detection with ViTs. \myacro\ applies joint, hierarchical pruning to image tokens and 3D queries, replacing hard drops with filter-and-store buffers that allow selective reactivation. Relevance is supervised by a future interaction corridor, and nuScenes-R evaluates accuracy specifically on planning-critical agents. On nuScenes, \myacro\ reduces latency by up to \(3\times\) with marginal accuracy loss; at aggressive sparsity (\myacro-1/10) it reaches \emph{real-time} throughput ($\sim$18 FPS with ViT-B) while achieving state-of-the-art performance among methods with similar latency. Since nuScenes-R depends on corridor hyperparameters \((H,d_{\min})\), future work includes tighter coupling of relevance with planning, LiDAR fusion, and closed-loop evaluation.

\clearpage

{
    \small
    \bibliographystyle{ieeenat_fullname}
    \bibliography{main}
}

\input{suppl}

\end{document}

%% file: preamble.tex
\usepackage{microtype}

\renewcommand{\paragraph}[1]{\vspace{.5em}\noindent\textbf{#1.}}

\definecolor{cvprblue}{rgb}{0.21,0.49,0.74}
\usepackage[pagebackref,breaklinks,colorlinks,allcolors=cvprblue]{hyperref}
\usepackage{bbm}
\newcommand{\myacro}{SToRe3D}
\definecolor{highlight}{RGB}{230,230,230}
\usepackage{multirow}

%% file: suppl.tex
\clearpage
\setcounter{page}{1}
\maketitlesupplementary

\section{Implementation Details}
\label{sec:supp_imp}

We follow standard multi-view 3D detection settings on nuScenes using 6 cameras, synchronized frames, and known camera intrinsics and extrinsics. Backbones are ViT-based, and we evaluate both medium and large variants. There is no encoder after the backbone, and the decoder follows a DETR-style design with multi-scale features. We measure end-to-end latency at batch size 1 with the standard PyTorch profiler on a single RTX3090 GPU.

The ViT-L and ViT-B backbones follow EVA-02~\cite{fang2024eva}, which is used for pretraining. The ViT-L backbone has 1024 embedding channels and 24 transformer layers with a mix of global and windowed attention, 16 attention heads, and a 0.3 drop path rate. The ViT-B backbone has 768 embedding channels and 12 transformer layers with a mix of global and windowed attention, 12 attention heads, and a 0.1 drop path rate. Both backbones use a patch size of 16. The nominal window size is 16 for ViT-L and 14 for ViT-B. 

The detection head uses 6 decoder layers with 256-dimensional query embeddings. In Table~\ref{tab:supp_hyp}, we list training hyperparameters and backbone configuration for all experiments, following EVA-02~\cite{fang2024eva} and StreamPETR~\cite{wang2023exploring} where possible. Hyperparameter and design choices specific to \myacro\ are described in the main text.

\begin{table}[!htb]
    \centering
    \begin{tabular}{l|l|l}
    \toprule
    \textbf{Base Config} & \multicolumn{2}{l}{\textbf{Value}}  \\ \midrule
    Learning rate & \multicolumn{2}{l}{4e-4} \\
    Learning rate schedule & \multicolumn{2}{l}{cosine decay} \\
    Optimizer & \multicolumn{2}{l}{AdamW} \\
    Weight decay & \multicolumn{2}{l}{0.01} \\
    Batch size & \multicolumn{2}{l}{16} \\
    Training epochs & \multicolumn{2}{l}{24} \\
    Max detections & \multicolumn{2}{l}{300} \\ \midrule
    \textbf{ViT Config} & \textbf{ViT-L} & \textbf{ViT-B}  \\ \midrule
    Checkpoint & \multicolumn{2}{l}{EVA-02~\cite{fang2024eva} Objects365} \\
    Patch size & 16 & 16 \\
    Embedding dimension & 1024 & 768 \\
    Window attention layers & 16 & 6 \\
    Global attention layers & 8 & 6 \\
    Attention heads & 16 & 12 \\
    Window size & 16 & 14 \\
    Drop path & 0.3 & 0.1 \\
    \end{tabular}
    \caption{Model configuration and training hyperparameters used in all experiments.}
    \label{tab:supp_hyp}
\end{table}

Pruning in the backbone occurs in the layer preceding a subset of global-attention blocks. For ViT-L with global attention at layers $[3, 6, 9, 12, 15, 18, 21, 24]$, pruning is applied before the later global layers. For ViT-B with global attention at layers $[2, 4, 6, 8, 10, 12]$, pruning is similarly applied before a subset of these layers. 

The pruning schedule for the number of tokens kept in each layer follows a quadratic schedule,
\begin{equation}
    \textrm{LKR}_l = \textrm{TKR} + (1-\textrm{TKR})(1-l/L)^2,
\end{equation}
where the layer keep ratio (LKR) is the fraction of tokens processed at layer $l$, the total keep ratio (TKR) is the final target fraction of tokens processed across the backbone, and $L$ is the total number of layers. Table~\ref{tab:pruning} lists the resulting per-layer keep ratios and the mean keep ratio (MKR), which approximates the average fraction of tokens processed through the backbone.
\begin{table}[!htb]
\centering
\resizebox{1\linewidth}{!}{
\begin{tabular}{l|c|c|cccc}
\toprule
\textbf{ViT-L Models}   & \textbf{TKR}  & \textbf{MKR}  & \textbf{LKR$_6$}    &  \textbf{LKR$_8$}    &  \textbf{LKR$_{10}$}    &  \textbf{LKR$_{12}$}    \\ \midrule
ToC3D-fast        & 0.5  & 0.70 & 0.70 & 0.50 & 0.50 & 0.50 \\
\myacro-1/2       & 0.5  & 0.76 & 0.63 & 0.53 & 0.50 & 1.00 \\ \midrule
ToC3D-faster      & 0.3  & 0.58 & 0.50 & 0.40 & 0.30 & 0.30 \\
\myacro-1/3       & 0.3  & 0.67 & 0.48 & 0.32 & 0.30 & 1.00 \\ \midrule
\myacro-1/10      & 0.1  & 0.58 & 0.33 & 0.16 & 0.10 & 1.00 \\ \midrule
\textbf{ViT-B Models}   & \textbf{TKR}  & \textbf{MKR}  & \textbf{LKR$_9$}    & \textbf{LKR$_{15}$}     &  \textbf{LKR$_{21}$}    & \textbf{LKR$_{24}$} \\ \midrule
ToC3D-fast        & 0.5 & 0.70 & 0.70 & 0.50 & 0.50 & 0.50 \\
\myacro-1/2      & 0.5 & 0.74 & 0.63 & 0.53 & 0.50 & 1.00 \\ \midrule
ToC3D-faster      & 0.3 & 0.58 & 0.50 & 0.40 & 0.30 & 0.30 \\
\myacro-1/3      & 0.3 & 0.64 & 0.48 & 0.34 & 0.30 & 1.00 \\ \midrule
\myacro-1/10     & 0.1 & 0.53 & 0.33 & 0.16 & 0.10 & 1.00 \\
\bottomrule
\end{tabular}
}
\caption{Token pruning schedules for \myacro\ model variants and ToC3D~\cite{zhang2024make}. TKR is the total keep ratio, MKR is the mean keep ratio across layers, and LKR$_\ell$ is the layer keep ratio at selected global-attention layers.}
\label{tab:pruning}
\end{table}



\section{Profiling Analysis}
\label{sec:supp_prof}

\begin{figure}[!ht]
    \centering
    \includegraphics[width=\linewidth]{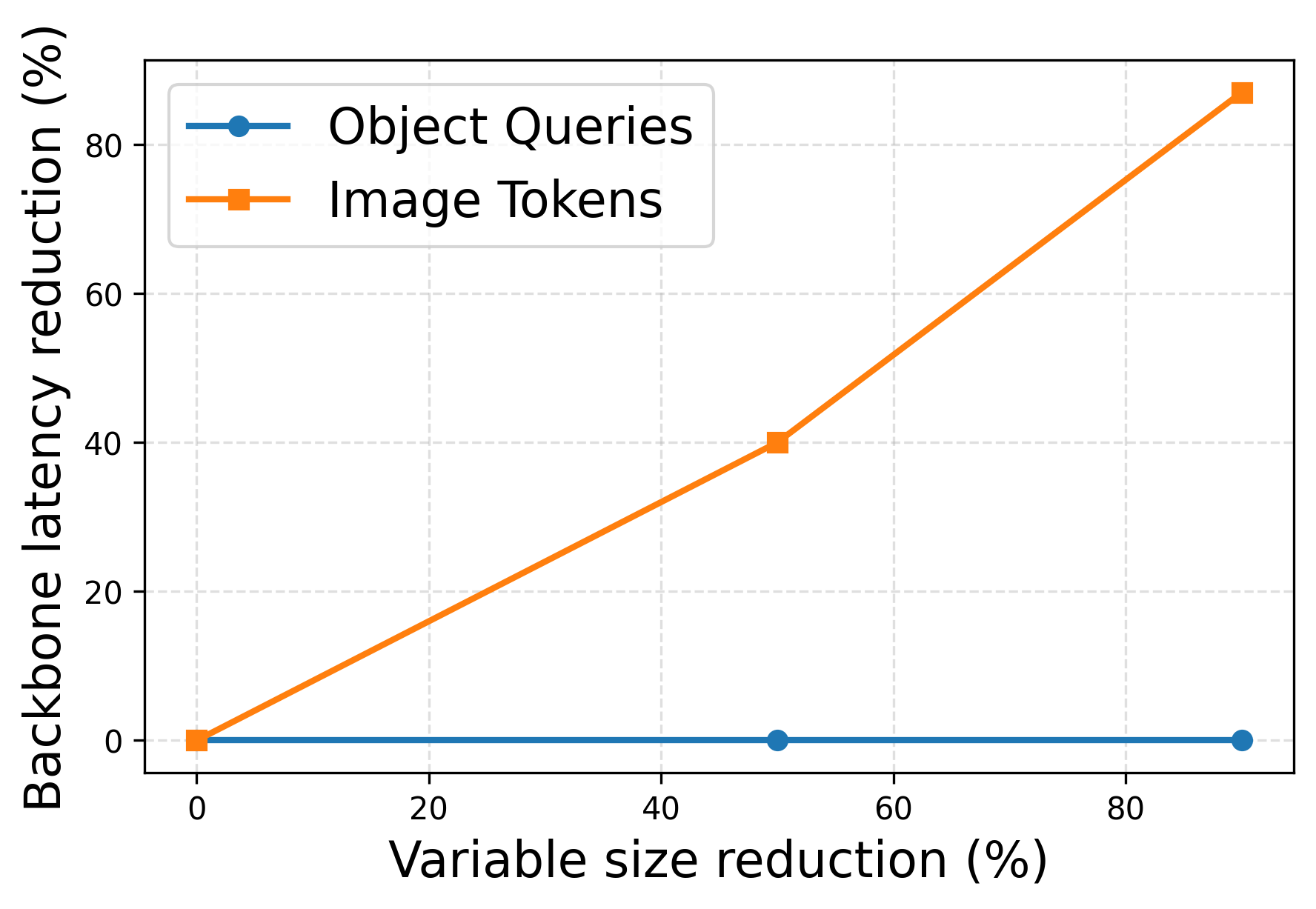}
    \includegraphics[width=\linewidth]{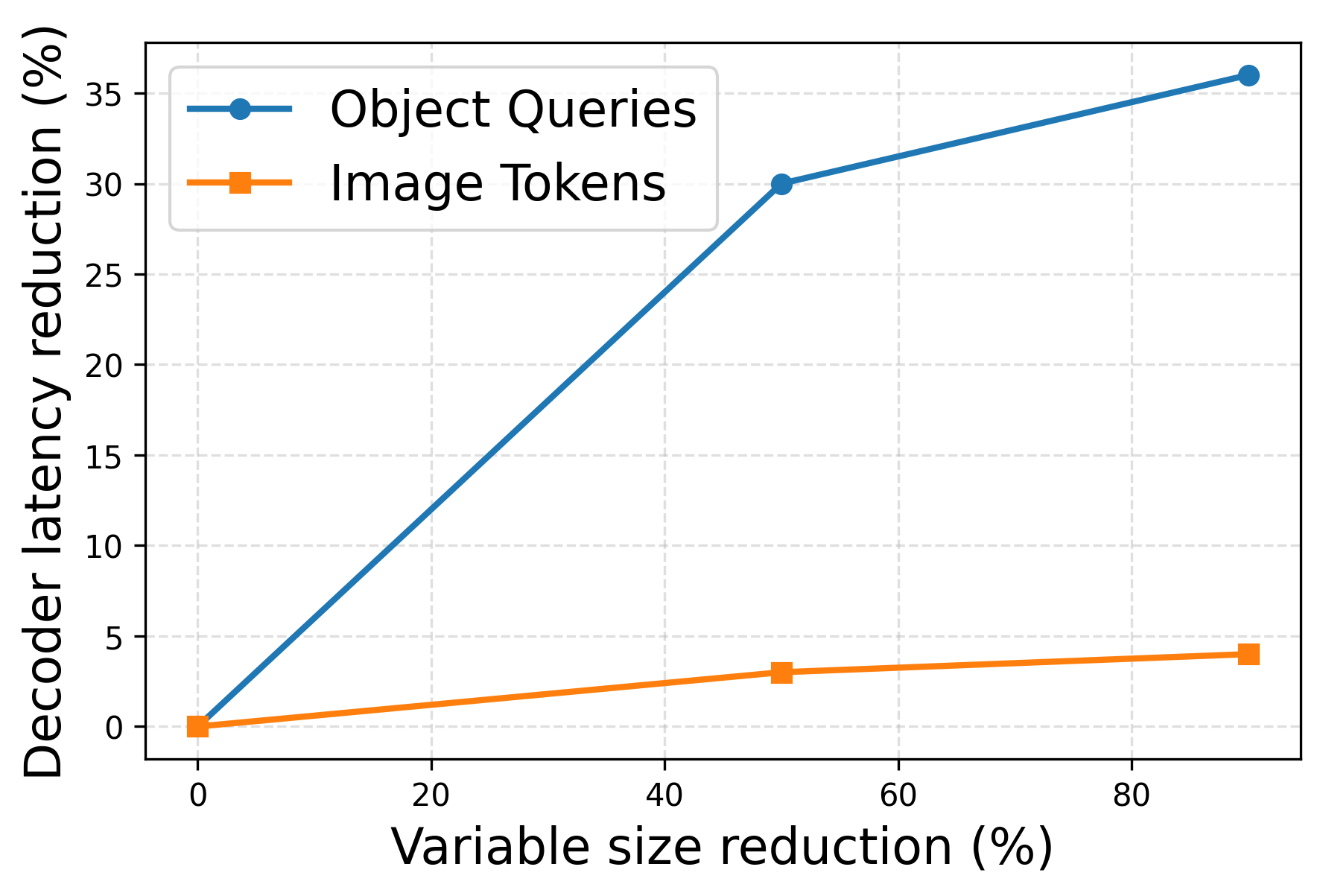}
    \caption{Latency sensitivity of the ViT backbone and the detection decoder when varying the number of image tokens and object queries independently. The backbone is almost entirely driven by the number of image tokens, while the decoder is much more sensitive to the number of object queries.}
    \label{fig:latency}
\end{figure}

We now analyze latency sensitivity to image tokens and object queries in isolation. Starting from the dense baseline with 900 object queries and 127{,}500 image tokens across all camera views, we systematically subsample each axis. We reduce the number of object queries by 50\% and 90\%, and measure the resulting latency of the ViT-L backbone and the detection decoder. We then repeat the experiment by reducing the number of image tokens by 50\% and 90\% while keeping the number of queries fixed. Figure~\ref{fig:latency} shows the measured latencies.

From these experiments, we estimate simple sensitivities $\Delta t / \Delta i$ and $\Delta t / \Delta o$, where $t$ is the latency, $i$ is the number of image tokens, and $o$ is the number of object queries. For the decoder, the sensitivity to image tokens is modest, with $\Delta t / \Delta i \approx 7\%$, while the sensitivity to object queries is much higher, with $\Delta t / \Delta o \approx 50\%$. For the ViT-L backbone, the sensitivity to image tokens is dominant, with $\Delta t / \Delta i \approx 95\%$, and the sensitivity to object queries is effectively zero, since queries are not used in the backbone.

These results support our design choice. Pruning image tokens primarily reduces backbone latency, while pruning object queries primarily reduces decoder latency. Joint sparsity over both axes is therefore necessary to reach the strongest latency improvements.

\section{Additional Metrics}
\label{sec:supp_metrics}

\begin{figure}[!ht]
    \centering
    \includegraphics[width=\linewidth]{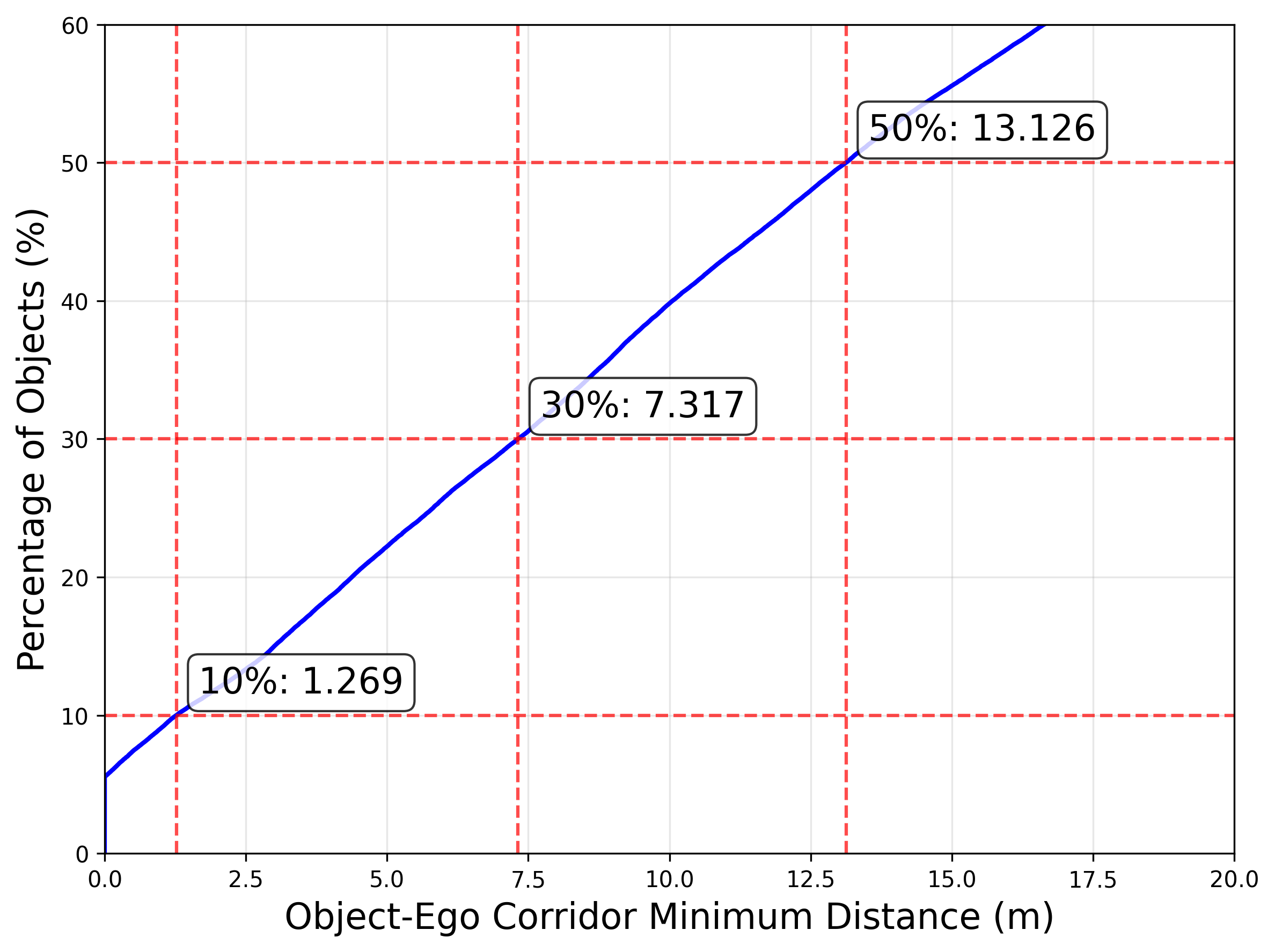}
    \caption{Cumulative distribution of closest ego–agent distances on nuScenes, used to select the relevance buffer for our nuScenes-R benchmark.}
    \label{fig:dist_cdf}
\end{figure}

For the planning-relevance metrics introduced in the main paper, an agent is labeled relevant if the closest distance $d_C$ between its swept corridor and the ego vehicle's swept corridor is below a buffer threshold $d_{RM}$. To choose $d_{RM}$, we first compute the empirical distribution of the closest ego–agent distances $d_C$ across the entire nuScenes dataset. The cumulative distribution is shown in Figure~\ref{fig:dist_cdf}. We select $d_{RM}$ as the 10th percentile of this distribution, which corresponds to approximately 10\% of agents being labeled relevant. This yields $d_{RM} = 1.2$ m over a 5-second horizon, which in turn leads to roughly 3 relevant agents per frame on average and a maximum of 31 relevant agents.


\section{Additional Qualitative Results}
\label{sec:supp_qual}

\begin{figure*}[!ht]
    \centering
    \begin{subfigure}[b]{\textwidth}
        \includegraphics[width=0.49\linewidth]{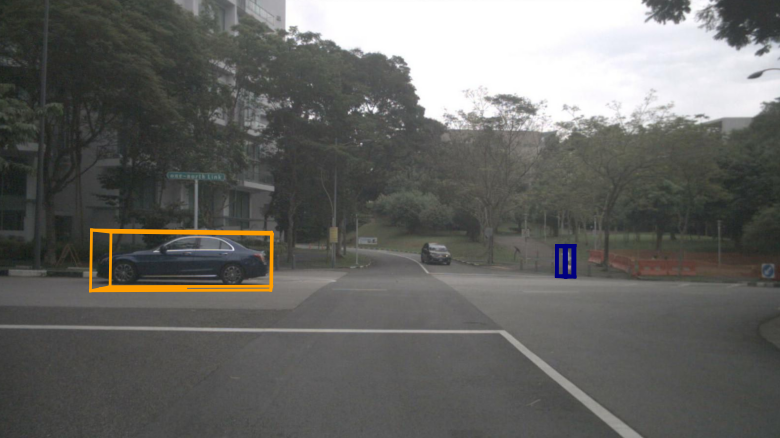}
        \hfill
        \includegraphics[width=0.49\linewidth]{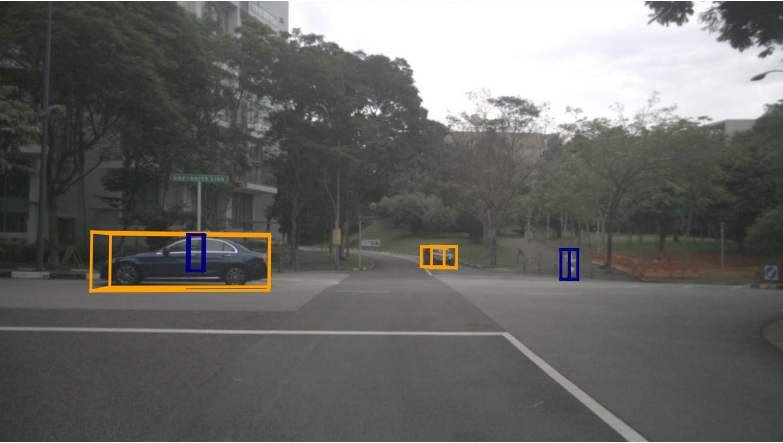}
        \caption{Front view of an oncoming car that is missed by the baseline detector but correctly detected by SToRe3D.}
        \label{fig:qual1a}
    \end{subfigure}
    \\
    \begin{subfigure}[b]{\textwidth}
        \includegraphics[width=0.49\linewidth]{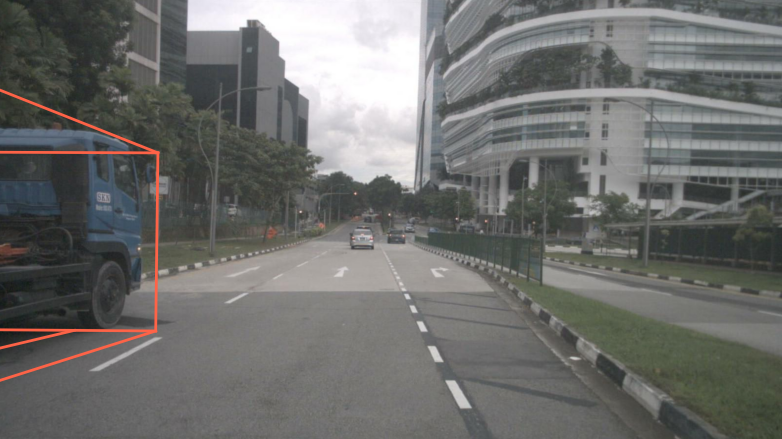}
        \hfill
        \includegraphics[width=0.49\linewidth]{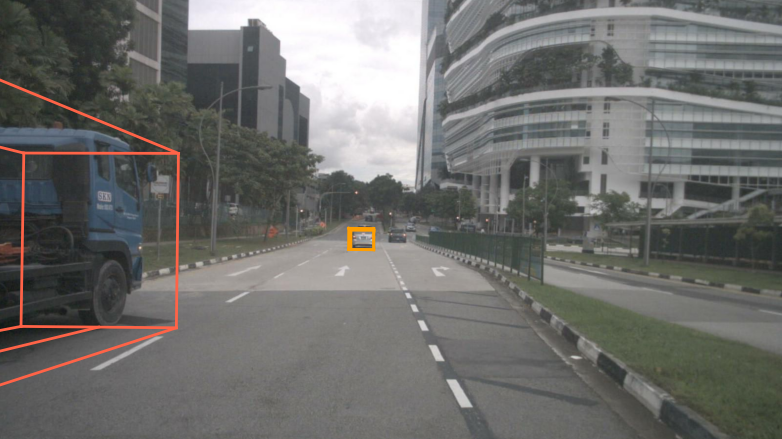}
        \caption{Front view of a lead vehicle ahead that is missed by the baseline but detected by SToRe3D.}
        \label{fig:qual2a}
    \end{subfigure}
    \\
    \begin{subfigure}[b]{\textwidth}
        \includegraphics[width=0.49\linewidth]{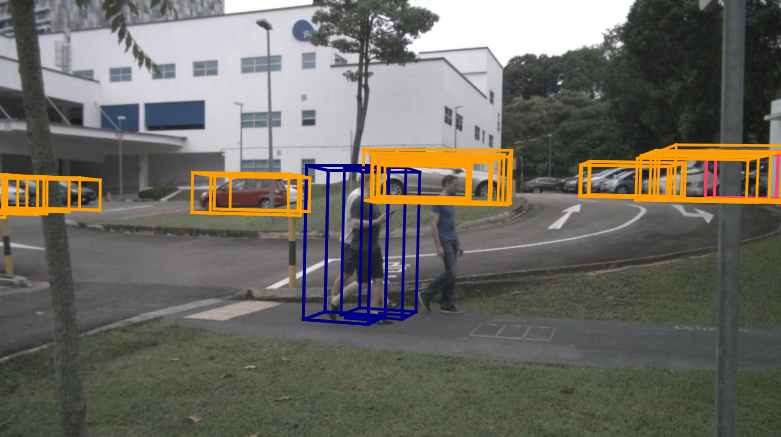}
        \hfill
        \includegraphics[width=0.49\linewidth]{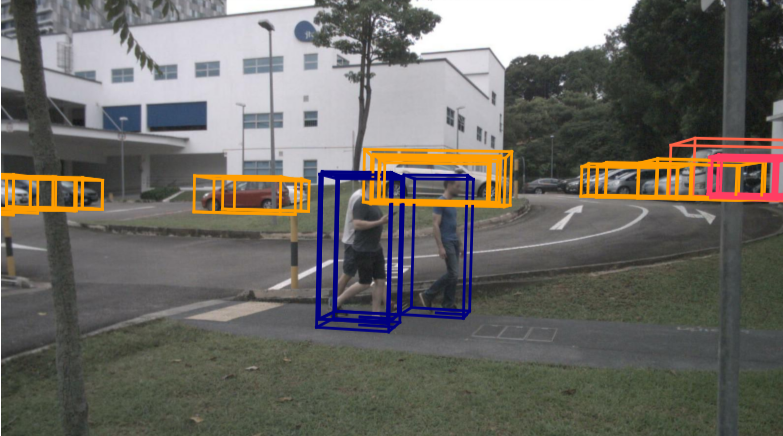}
        \caption{Left side view of a nearby pedestrian that is missed by the baseline but detected by SToRe3D.}
        \label{fig:qual2b}
    \end{subfigure}
    \caption{Additional qualitative examples of false negatives for the baseline StreamPETR-R50 (left in each pair), where SToRe3D-1/10-ViT-B at a similar latency (right in each pair) correctly detects the planning-critical objects.}
    \label{fig:supp_qual}
\end{figure*}

Figure~\ref{fig:supp_qual} provides additional qualitative comparisons between \myacro\ and the baseline StreamPETR model. We highlight three representative cases where highly relevant objects are missed by the baseline ResNet-50 model but detected by our similar-latency variant SToRe3D-1/10-ViT-B.

In all three scenes, the highlighted objects interact with the ego vehicle and have predicted future trajectories that pass close to the ego path. Examples include a car in the oncoming lane at an intersection (top row), a car stopped ahead near an upcoming intersection (middle row), and a nearby pedestrian about to cross the road (bottom row). These cases illustrate that the higher-capacity ViT backbone with relevance-aware pruning detects planning-critical objects that a lower-capacity dense backbone misses, while operating at comparable latency.


\section{Limitations and Future Work}

While \myacro\ achieves strong accuracy–latency trade-offs, it has several limitations. First, our evaluation is restricted to the nuScenes dataset and camera-only multi-view 3D detection. The relevance heads and pruning schedules are tuned for this setting. Performance and sparsity behavior on other datasets, sensing setups (for example, LiDAR or radar fusion), and driving domains are not evaluated.

Second, the approach still relies on relatively heavy ViT backbones and multi-scale DETR-style decoding. Although relevance-aware sparsity yields substantial speedups, absolute latency and memory usage may remain challenging for some embedded or low-power deployments, especially at higher input resolutions.

Finally, the nuScenes-R protocol for planning-critical evaluation uses privileged future information to define interaction corridors. This is reasonable for offline benchmarking, but the corridor parameters and labeling procedure may not perfectly match the objectives of a downstream planner in a closed-loop system. In addition, we focus only on open-loop detection and do not assess downstream effects on full autonomous driving stacks in closed-loop simulation or real-world testing. Studying how relevance-aware sparsity interacts with planning and control policies, along with joint training of SToRe3D for end-to-end driving are important direction for future work.